\documentclass{llncs}
\usepackage{graphicx}
\usepackage{subfig}
\begin{document}

\title{Classification of breast cancer histology images using transfer learning}

\author{Sulaiman Vesal$^1$, Nishant Ravikumar$^1$, AmirAbbas Davari$^1$, Stephan Ellmann$^2$, Andreas Maier$^1$}
    
\institute{$^1$Pattern Recognition Lab, Friedrich-Alexander-Universit\"at Erlangen-N\"urnberg\{sulaiman.vesal@fau.de\}
      \\$^2$Radiologisches Institut, Universit\"atsklinikum Erlangen, Germany}
    
    \maketitle
    
\begin{abstract}
        
        Breast cancer is one of the leading causes of mortality in women. Early detection and treatment are imperative for improving survival rates, which have steadily increased in recent years as a result of more sophisticated computer-aided-diagnosis (CAD) systems. A critical component of breast cancer diagnosis relies on histopathology, a laborious and highly subjective process. Consequently, CAD systems are essential to reduce inter-rater variability and supplement the analyses conducted by specialists. In this paper, a transfer-learning based approach is proposed, for the task of breast histology image classification into four tissue sub-types, namely, normal, benign, \textit{in situ} carcinoma and invasive carcinoma. The histology images, provided as part of the BACH 2018 grand challenge, were first normalized to correct for color variations resulting from inconsistencies during slide preparation. Subsequently, image patches were extracted and used to fine-tune Google`s Inception-V3  and ResNet50 convolutional neural networks (CNNs), both pre-trained on the ImageNet database, enabling them to learn domain-specific features, necessary to classify the histology images. The ResNet50 network (based on residual learning) achieved a test classification accuracy of 97.50\%  for four classes, outperforming the Inception-V3 network which achieved an accuracy of 91.25\%.
        
    \end{abstract}
    
\section{Introduction}\label{sec:Introduction}
    
    According to a recent report published by the American Cancer Society, breast cancer is the most prevalent form of cancer in women, in the USA. In 2017 alone, studies indicate that approximately 252,000 new cases of invasive breast cancer and 63,000 cases of \textit{in situ} breast cancer are expected to be diagnosed, with 40,000 breast cancer-related deaths expected to occur \cite{desantis17}. Consequently, there is a real need for early diagnosis and treatment, in order to reduce morbidity rates and improve patients' quality of life. Histopathology remains crucial to the diagnostic process and the gold standard for differentiating between benign and malignant tissue, and distinguishing between patients suffering from \textit{in situ} and invasive carcinoma \cite{Xu2017}. Diagnosis and identification of breast cancer sub-types typically involve collection of tissue biopsies from masses identified using mammography or ultrasound imaging, followed by histological analysis. Tissue samples are usually stained with Hematoxylin and Eosin (H\&E) and subsequently, visually assessed by pathologists using light microscopy. Visual assessment of tissue microstructure and the overall organization of nuclei in histology images is time-consuming and can be highly subjective, due to the complex nature of the visible structures. Consequently, automatic computer-aided-diagnosis systems are essential to reduce the workload of specialists by improving diagnostic efficiency, and to reduce subjectivity in disease classification.

    Classification of histology images into cancer sub-types and metastases detection in whole-slide images are challenging tasks. Numerous studies have proposed automated approaches to address the same in recent years. Kothari et al. \cite{Kothari2013} examined the utility of biologically interpretable shape-based features for classification of histological renal tumor images. They extracted shape-based features that captured the distribution of tissue structures in each image and employed these features within a multi-class classification model. Doyle et al. \cite{doyle08} proposed an automated framework for distinguishing between low and high grades of breast cancer, from H\&E-stained histology images. They employed a large number of image-derived features together with spectral clustering to reduce the dimensionality of the feature space. The reduced feature set was subsequently used to train a support vector machine classifier to distinguish between cancerous and non-cancerous images, and low and high grades of breast cancer. Wang et al. \cite{wang16} proposed an award-winning (at the International Symposium on Biomedical Imaging) deep learning framework for whole-slide classification and cancer metastases detection in breast sentinel lymph node images. In a recent study \cite{10.1371/journal.pone.0177544}, the authors proposed a convolutional neural network (CNN) based approach to classifying H\&E-stained breast histology images into four tissue classes, namely, healthy, benign, \textit{in situ} carcinoma and invasive carcinoma, with a limited number of training samples. The features extracted by the CNN were used for training a Support Vector Machine classifier. Accuracies of 77.8\% for four class classification and 83.3\% for carcinoma/non-carcinoma classification were achieved. In this study, we investigate the efficacy of transfer-learning for the task of image-wise classification of H\&E-stained breast cancer histology images and examine the classification performance of the pre-trained Inception-V3 \cite{szegedy16} and ResNet50 \cite{he2016} networks, on the BACH 2018 challenge data set.

    \section{Methods}\label{sec:Materials and Method}
    
    The data set used in this study was provided as part of BACH 2018 grand challenge\footnote{https://iciar2018-challenge.grand-challenge.org/home/}, comprising H\&E-stained breast histology microscopy images. The images are high-resolution (2040 $\times$ 1536 pixels), uncompressed, and annotated as normal, benign, \textit{in situ} carcinoma or invasive carcinoma, as per the predominant tissue type visible in each image. The annotation was performed by two medical experts and images with disagreements were discarded. All images were digitized using the same acquisition conditions, with a magnification of 200$\times$. 
    
    The data set comprises 400 images (100 samples in each class), with a pixel scale of 0.42 $\mu$m $\times$ 0.42 $\mu$m. The provided images were selected such that pathology classification could be objectively determined from the tissue structure and organization visible. The data set was partitioned into training (60 samples), validation (20 samples), and test (20 samples) sets, by selecting images at random for each class independently.

    \subsection{Stain Normalization}
    
    A common problem with histological image analysis is substantial variation in color between images due to differences in color responses of slide scanners, raw materials and manufacturing techniques of stain vendors, and staining protocols. Consequently, stain normalization is essential as a pre-processing step, prior to conducting any analyses using histology images. Various strategies \cite{7164042}, \cite{5193250} have been proposed for stain normalization in histological images. In this paper, we used the approach proposed by Reinhard et al. \cite{946629} which matches the statistics of color histograms of a source and target image, following transformation of the RGB images to the de-correlated LAB color space. Here, the mean and standard deviation of each channel in the source image is matched to that of the target by means of a set of linear transforms in the LAB color space. Histogram matching techniques assume that the proportions of stained tissue components for each staining agent are similar across the images being normalized. Fig. 2 illustrates the effect of stain normalization on a few samples from the breast cancer histology image data set using the method proposed in \cite{946629}.
    
    \begin{figure}[ht]
        \centering
        \includegraphics[width=0.99\textwidth]{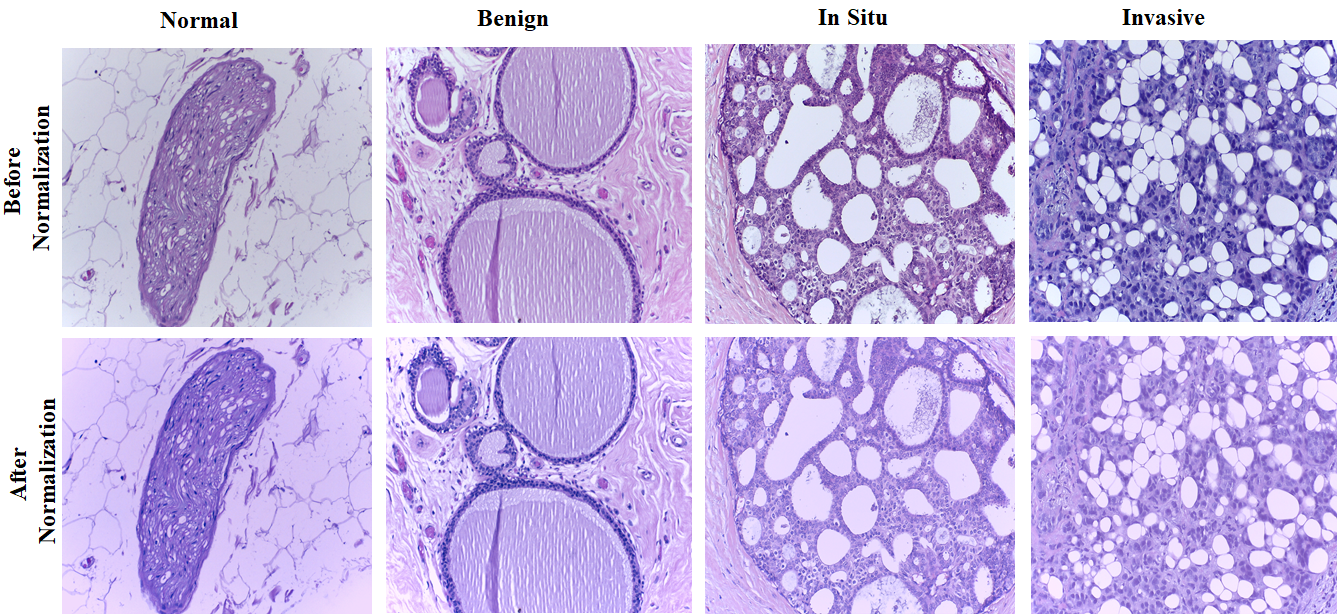}
        \caption{Examples of histology images from each class before (top row) and after (bottom row) stain normalization.}
        \label{fig1}
    \end{figure}
    
    \subsection{Pre-processing}
    
    Deep learning approaches are heavily dependent on the volume of training data available, with models of higher complexity requiring more data to generalize well and avoid over-fitting to the training samples. A common challenge in the medical domain is a lack of sufficient data, as was the case with the BACH 2018 challenge. Additionally, the breast histology images provided in the challenge data set are very large in size, spanning 2040 $\times$ 1536 pixels. In order to address the issues of limited data and large image sizes, we extracted patches from each image and augmented the data set using a variety of rigid transformations, thereby increasing the number of training samples. Image-wise classification into tissue/cancer sub-types requires learning features describing overall tissue architecture and localized organization of nuclei. Consequently, we chose to extract patches of size 512 $\times$ 512 pixels from each image, while ensuring 50\% overlap between patches (similar to \cite{10.1371/journal.pone.0177544}), as there was no guarantee that smaller patches would contain information relevant to the class assigned to the whole image. This resulted in the extraction of 35 patches from each image and a final data set comprising 14,000 patches.
    
    Additionally, to enrich the training set we augmented the data by applying varying degrees of rotation, and flipping the extracted patches. This mode of data augmentation emulates a real-world scenario as there is no fixed orientation adopted by pathologists when analyzing histology slides/images. Such a patch extraction and dataset augmentation approach have been used previously for an identical classification problem \cite{10.1371/journal.pone.0177544}. The training data was augmented by flipping the extracted patches along their horizontal and vertical edges and rotating them by 90, 180, 270 degrees. Thus, each patch was transformed to create 5 additional, unique patches resulting in a total of 67,200 training and validation patches from the original 320 training images. The label for each patch was inherited from the class assigned to the original image. The remaining `unseen' 80 images were used as test data, to evaluate the classification accuracy of the methods investigated.
    
    \subsection{Pre-trained CNN Architectures}
    The application of CNNs pre-trained on large annotated image databases, such as ImageNet for example, to images from different modalities/domains, for various classification tasks, is referred to as transfer learning. Pre-trained CNNs can be fine-tuned on medical image data sets, enabling large networks to converge quicker and learn domain-/task-specific features. Fine-tuning pre-trained CNNs is crucial for their re-usability \cite{Yosinski:2014:TFD:2969033.2969197}. With such an approach, the original network architecture is maintained and the pre-trained weights are used to initialize the network. The initialized weights are subsequently updated during the fine-tuning process, enabling the network to learn features specific to the task of interest. Recently, numerous studies have demonstrated that fine-tuning is effective and efficient for a variety of classification tasks in the medical domain \cite{shin16}. In this study, we investigate two well known pre-trained CNN architectures, namely, Google`s Inception-V3 \cite{szegedy16} and deep residual convolutional (ResNet50) network \cite{he2016}, which are fine-tuned to learn domain and modality specific features for classifying breast histology images. ResNet50 is based on a residual learning framework where, layers within a network are reformulated to learn a residual mapping rather than the desired unknown mapping between the inputs and outputs. Such a network is easier to optimize and consequently, enables training of deeper networks, which correspondingly leads to an overall improvement in network capacity and performance. A recent study showed that Google`s Inception-V3 network, pre-trained on ImageNet and fine-tuned using images of skin lesions, achieved very high accuracy for skin cancer classification, comparable to that of numerous dermatologists \cite{esteva17}. The Inception-V3 network employs factorized inception modules, allowing the network to choose suitable kernel sizes for the convolution layers. This enables the network to learn both low-level features with small convolutions and high-level features with larger ones.

\begin{figure}[ht]
        \centering
        \includegraphics[width=0.99\textwidth]{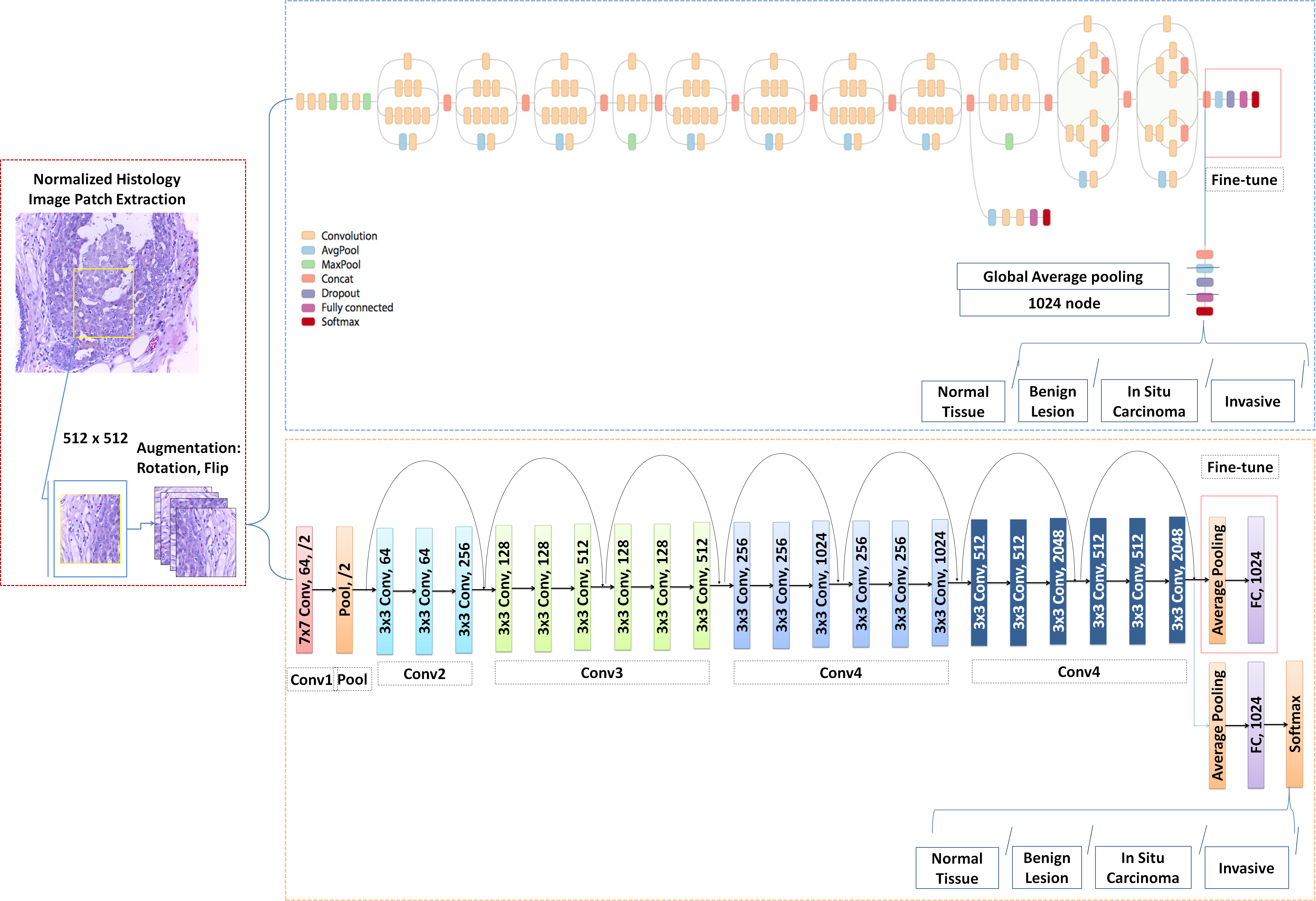}
        \caption{Breast histology image classification workflow by fine-tuning Google`s Inception-V3 and ResNet50 network architectures. The block on the left represents the pre-processing steps and the blocks on the right depict the Inception-V3 (top) and ResNet50 (bottom) network architectures.}
        \label{fig1}
    \end{figure}
    
        The dataset was pre-processed as described in the previous section and used to fine-tune Google`s Inception-V3 and ResNet50 networks (both pre-trained on ImageNet). While such a transfer learning approach has been adopted for a variety of classification and detection tasks in medical images, few studies have employed the same for breast cancer histology image classification. Fig. 2 describes our proposed workflow for the Inception-V3 and ResNet50 network architectures. The original Inception network is modified by replacing the last 5 layers with an average global pooling layer, 1 fully connected layer, and a softmax classifier. The latter outputs probabilities for each of the four classes of interest, for each patch, fed as input to the network during the fine-tuning process. The stochastic gradient descent optimizer with momentum was employed to train the Inception-V3 network, with a batch size of 32 for both training and validation. A learning rate and Nesterov momentum of 0.0001 and 0.9, respectively, were found to be suitable. The network stopped learning after 100 epochs. The same fine-tuning approach was applied to the ResNet50 network with identical optimization parameters. Model performance was measured by first classifying several patches extracted from each unseen test image, and then combining the classification results of all patches through a majority voting process, to obtain the final class label for each image.

\section{Results}
    
    We conducted several experiments on the challenge data set to evaluate the classification performance of the networks investigated. First, the overall prediction accuracy of the networks was assessed as the ratio between the number of images classified correctly and the total number images evaluated. Patch-wise and image-wise classification accuracy are presented in Table 1 for both ResNet50 and Inception-V3 networks. Patch-wise classification accuracy of ResNet50 for the validation and test sets were 90.68\% and 94.50\%, respectively. The Inception-V3 network on the other hand achieved patch-wise classification accuracies of 87.14\% and 86.57\% for the validation and test sets, respectively. As discussed previously, whole image classification was achieved using a majority voting process, based on the patch-wise class labels estimated using each network. ResNet50 achieved whole-image classification accuracies of 89.58\% and 97.50\%, for the validation and test sets, respectively. Meanwhile, the Inception-V3 network achieved classification accuracies of 87.03\% and 91.25\% for the validation and test sets, respectively. Overall, ResNet50 consistently outperformed the Inception-V3 network, achieving higher patch-wise and image-wise classification accuracy, for both the validation and test data.
  
    \begin{table}[ht]
\centering
\caption{Patch-wise and image-wise classification accuracy(\%) for the ResNet50 and Inception-V3 networks.}
        \label{0000-tab-schriften}
        \begin{tabular*}{\textwidth}{l@{\extracolsep\fill}llll} \\ \hline
Model        & \multicolumn{2}{c}{Patch-Wise} & \multicolumn{2}{c}{Image-Wise} \\ \hline
        & Validation Set(\%)    & Test Set(\%)   & Validation Set(\%)    & Test Set(\%)   \\ \hline
ResNet50 & 90.68              &94.03          &    89.58             & 97.50         \\ \hline
Inception-V3     & 87.14                & 86.57        & 87.03                & 91.25         \\ \hline
\end{tabular*}
\end{table}

\begin{figure}[!htp]
  % Maximum length
 \subfloat{\includegraphics[width=0.99\linewidth]{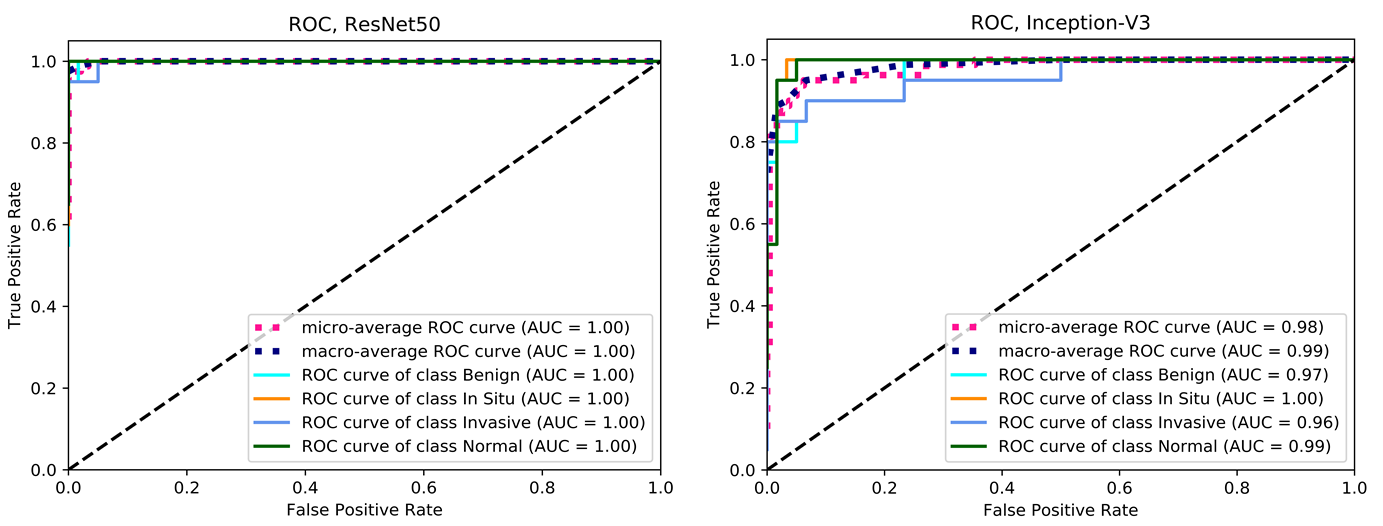}}\hfill

  \caption{ROC curves for unseen test set using Google`s Inception-V3 and ResNet50 fine-tuned architectures.}
\end{figure}

We also computed the receiver operating characteristic (ROC) curves for each network \cite{Xu2017}, depicted in Fig. 3. ROC curves plot the true positive rate (TPR) versus the false positive rate (FPR) at different threshold settings. TPR also known as sensitivity, represents the proportion of correctly classified samples and FPR, also known as fall-out, represents the proportion of incorrectly classified samples. Thus classification accuracy was measured as the area under the ROC curve (AUC), with an area of 1 representing perfect classification on the test set. We assessed network performance for each class individually by computing their ROCs and calculated their corresponding AUCs (presented in Fig. 3). The overall specificity and sensitivity of ResNet50 is approximately 99.9\%  and that of Inception-V3 is 98\%.

\section{Conclusions}
    A transfer learning-based approach for classification of H\&E-stained histological breast cancer images is presented in this study. The network learns features using Google`s Inception-V3 and residual network (ResNet50) architectures, which have been pre-trained on ImageNet. The data set of images provided for the BACH 2018 grand challenge are classified into four tissue classes, namely, normal, benign, \textit{in situ} carcinoma and invasive carcinoma. We trained both networks using 60\% of the data set, validated on 20\% and evaluated their performance on the remaining 20\% of images. The proposed transfer-learning approach is simple, effective and efficient for automatic classification of breast cancer histology images. The investigated networks successfully transferred ImageNet knowledge encoded as convolutional features to the problem of histology image classification, in the presence of limited training data. The residual network (ResNet50) investigated, outperformed Google`s Inception-V3 network consistently, in terms of classification accuracy. The presented work demonstrates the applicability and powerful classification capacity of transfer learning approaches, for the automatic analysis of breast cancer histology images.
    
    %\section*{Acknowledgments}\label{sec:Acknowledgments}
    
    %Authors would like to thank YYYYY.
    
    \bibliographystyle{splncs}
    \bibliography{0000}

\end{document}